\documentclass[runningheads]{llncs}
\usepackage{graphicx}
\usepackage{booktabs}
\usepackage{subfigure}
\usepackage{dsfont}
\usepackage{amsmath}
\usepackage{amssymb}
\newcommand{\ptf}{predicted urban flow tensor}
\newcommand{\pr}{predicted periodic relation}
\newcommand{\name}{TERMCast}

\begin{document}
% \title{Urban Flow Forecasting via Long Term \\ Relation Modeling}
% \title{Relation Modeling: A New Perspective for Modeling Periodicity in Urban Flow Forecasting}
% \title{Joint Relation Modeling of Temporal Patterns in Urban Flow Forecasting}
% \title{Relation Modeling of Short- and Long-Term Patterns in Urban Flow Forecasting}
\title{TERMCast: Temporal Relation Modeling for Effective Urban Flow Forecasting}
% \title{TRAILER: \underline{Tra}nsformer-based T\underline{i}me-wise \underline{L}ong T\underline{e}rm \underline{R}elation Modeling for Citywide Crowd Flow Prediction}
%
\titlerunning{Temporal Relation Modeling for Effective Urban Flow Forecasting}
\author{Hao Xue \and
Flora D. Salim }
% \author{Anonymous Author(s)}
%
\authorrunning{H. Xue and F.D. Salim}
% \authorrunning{Anonymous Author}
% \institute{}
\institute{School of Computing Technologies, RMIT University, Melbourne, Australia \\
\email{\{hao.xue,flora.salim\}@rmit.edu.au}}
\maketitle              % typeset the header of the contribution
\begin{abstract}
Urban flow forecasting is a challenging task, given the inherent periodic characteristics of urban flow patterns.
To capture the periodicity, existing urban flow prediction approaches are often designed with \textit{closeness}, \textit{period}, and \textit{trend} components extracted from the urban flow sequence.
However, these three components are often considered separately in the prediction model.
These components have not been fully explored together and simultaneously incorporated in urban flow forecasting models.
We introduce a novel urban flow forecasting architecture, \name.
% To discover the periodicity and jointly model three components, a
A Transformer based long-term relation prediction module is explicitly designed to discover the periodicity and enable the three components to be jointly modeled
This module predicts the periodic relation which is then used to yield the \ptf.
To measure the consistency of the \pr\ vector and the relation vector inferred from the \ptf, we propose a consistency module. 
A consistency loss is introduced in the training process to further improve the prediction performance.
Through extensive experiments on three real-world datasets, we demonstrate that \name\ outperforms multiple state-of-the-art methods.
The effectiveness of each module in~\name\ has also been investigated.

\end{abstract}
\section{Introduction}
\begin{figure}
    \centering
    \includegraphics[width=.6\textwidth]{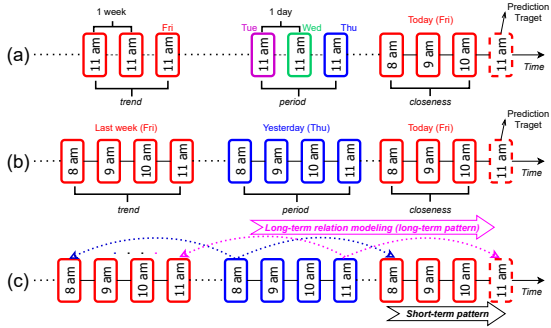}
    \caption{Illustration of the \textit{closeness}, \textit{period}, and \textit{trend} components. Figure (a) and (b) are two widely used construction methods of these components in the literature.
    In the proposed~\name, to predict urban flows at the prediction target, we focus on exploring the long-term relation among these components as shown in (c).}
    \label{fig:intro}
\end{figure}

Forecasting urban flows is essential for a wide range of applications from public safety control, urban planning to intelligent transportation systems.
For example, taxi and ride sharing companies are capable of providing better services with more accurate flow prediction.
From the user's perspective, based on the predicted urban flows, commuters and travelers are able to avoid traffic congestion and arrange driving routes.
Unlike other temporal sequence formats like language sentences and pedestrian trajectories, urban flow data has a unique inherent feature, that is, the periodicity, e.g., morning rush hours are more likely to occur during weekdays instead of weekends. 
To explore this periodicity feature, \textit{closeness}, \textit{period}, and \textit{trend} components are widely used as the input of a urban flow prediction model.
Two different settings of the three components are illustrated in Fig.~\ref{fig:intro} (a) and (b).
Fig.~\ref{fig:intro} (a) is widely used in existing methods~\cite{zhang2017deep,zonoozi2018periodic,zhang2018predicting,kang2019deep,lin2019deepstn} and Fig.~\ref{fig:intro} (b) is recently proposed by Jiang et al.~\cite{jiang2019vluc}.
In these settings, the \textit{closeness} component corresponds to the most recent observations, while the \textit{period} and \textit{trend} components reflect the daily and weekly periodicity, respectively.

Nevertheless, we notice that the relation of the three components is not well researched in the literature.
Existing methods always process these input components separately (Fig.~\ref{fig:intro}~(a) and (b)) to extract features of \textit{closeness}, \textit{period}, and \textit{trend}.
These features are then combined for urban flow prediction through a simple weighted fusion step such as ST-ResNet~\cite{zhang2017deep} and PCRN~\cite{zonoozi2018periodic} or network structures like the \textit{ResPlus} in DeepSTN+~\cite{lin2019deepstn} and the attention mechanism in VLUC-Net~\cite{jiang2019vluc}.
Although these methods have shown good abilities to predict urban flows, there are still unsolved questions such as what is the relation of each time interval among the \textit{closeness}, \textit{period}, and \textit{trend} component and how can these periodic relations facilitate the prediction?
For example, in Fig.~\ref{fig:intro} (c), if the relations at \textit{8 am}, \textit{9 am}, and \textit{10 am} (blue arrows) are modeled, can we predict the periodic relation at \textit{11 am} (pink arrows) and use this \pr\ to help the prediction of urban flow at \textit{11 am}?

Our research is motivated by the aforementioned questions. We propose a novel urban flow prediction network~\name\ in this paper.
Specifically, in the proposed~\name, a short-term prediction module is developed to explore the short-term pattern and generate an initial predicted flow.
This module handles the most up-to-date \textit{closeness} component which is a strong cue for the prediction target.
For better modeling the long-term pattern, rather than processing \textit{closeness}, \textit{period}, and \textit{trend} individually, we design a Transformer~\cite{vaswani2017attention} based long-term relation prediction module to model the periodic relation per time step among urban flows from the three components at the same time interval (e.g., 8 am, 9 am, and 10 am shown in Fig.~\ref{fig:intro} (c)) and predict the relation of our target time interval (e.g., 11 am).
This \pr\ and the initial flow from the short-term prediction module are then combined together to yield the \ptf.
So far, for the prediction target, there are two types of relations available: the \textit{\pr} generated by the long-term relation prediction module; and the \textit{inferred relation} that can be calculated based on the \ptf.
Intuitively, these two relations should be consistent with each other as they both belong to the same time interval.
To this end, we also design a prediction consistency module and propose a consistency loss to reflect such consistency in the training process of our network. 
In summary, our contributions are:
% \begin{itemize}
    (i) We propose a novel Transformer based long-term relation module to explore the relation among the \textit{closeness}, \textit{period}, and \textit{trend} components. It captures the long-term periodicity in the urban flow sequence and works together with the short-term prediction module to yield urban flow predictions. 
    (ii) We design a prediction consistency module to build a connection between the predicted urban flow tensor and the predicted periodic relation generated by the long-term relation module. 
    Furthermore, we propose a consistency loss upon the prediction consistency module to improve the accuracy of the prediction.
    (iii) We conduct extensive experiments on three publicly available real-world urban flow datasets. The results show that~\name\ achieves state-of-the-art performance.
% \end{itemize}

\section{Related Work}
The urban flow prediction problem and similar problems such as crowd flow prediction and taxi demand prediction have been the focus of researchers for a quite long time.
Based on the famous time series prediction model Auto Regressive Integrated Moving Average (ARIMA) and its variants such as Seasonal ARIMA, numerous traditional approaches including~\cite{williams2003modeling,kamarianakis2003forecasting,shekhar2007adaptive,li2012prediction,lippi2013short} have been designed and proposed.
In the last few years, deep learning based neural networks such as ResNet~\cite{he2016deep} and Transformer~\cite{vaswani2017attention} have been widely used in the areas of computer vision and neural language processing.
In the area of temporal sequence modeling, deep learning based methods have also been proposed and successfully applied to many time series prediction problems such as human mobility prediction~\cite{feng2018deepmove} and high-dimensional time series forecasting~\cite{sen2019think}.

\begin{figure*}[!t]
    \centering
    \includegraphics[width=0.9\textwidth]{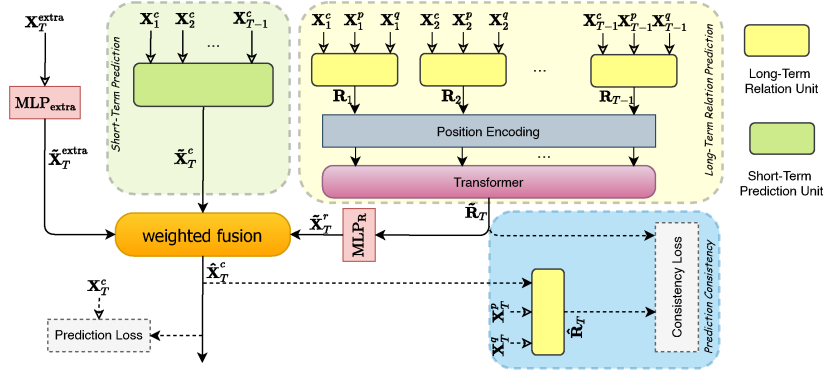}
    \caption{The architecture of the proposed~\name. Dash arrows indicate operations used in the training phase only.}
    \label{fig:framework}
\end{figure*}

More specifically, for urban flow prediction task, to model the periodic pattern of the urban flow sequence, Deep-ST~\cite{zhang2016dnn} firstly proposes to use \textit{closeness}, \textit{period}, and \textit{trend} three components (e.g., Fig.~\ref{fig:framework} (a)) to form the input instance of the prediction network.
Based on Deep-ST, ST-ResNet~\cite{zhang2017deep} employs convolution-based residual networks to further model the spatial dependencies in the city scale.
In ST-ResNet, three residual networks with the same structure are used to extract spatio-temporal features from three components respectively.
To yield predictions, the weighted fusion of these three features are then combined with the extracted extra features from an external component, such as weather conditions and events.
To capture spatial and temporal correlations, a pyramidal convolutional recurrent network is introduced in Periodic-CRN (PCRN)~\cite{zonoozi2018periodic}.
Yao et al.~\cite{yao2018deep} proposed a unified multi-view model that jointly considers the spatial, temporal, and semantic relations for flow prediction. 
STDN~\cite{yao2019revisiting} further improves the DMVST by proposing a periodically shifted attention mechanism to incorporate the long-term periodic information that is ignored in the DMVST.
Following the trend of fusing periodic representations (i.e., the \textit{period} and \textit{trend} components), more approaches are designed and proposed for the prediction of urban flow.
DeepSTN+~\cite{lin2019deepstn} proposes an architecture of \textit{ResPlus} unit, whereas VLUC-Net~\cite{jiang2019vluc} utilizes the attention mechanism in the fusion process.
However, these methods only consider and fuse periodic components after processing the three components separately.
This late fusion manner cannot fully explore the intrinsic relation in these components.
For example, the relation at \textit{8 am} (the blue dash arrow in Fig.~\ref{fig:intro} (c)) should be able to provide cues for the prediction at \textit{11 am} but it is overlooked by existing methods.

In summary, the proposed~\name\ differs from other methods in the aspect of modeling the periodicity, i.e. the long-term relation. 
We also introduce a novel consistency loss to connect the inferred relation and the \pr, which is different from the training loss functions of other urban flow prediction networks.
Our work is also different from~\cite{lai2018modeling} that uses CNN and RNN to extract short- and long-term patterns for multivariate time series forecasting.

\section{Preliminaries}
Following the widely-used grid-based urban flow definition from~\cite{zhang2017deep,jiang2019vluc}, we divide a city area into $H\times W$ disjoint regions based on the longitude and latitude.

\noindent\textbf{Inflow/Outflow~\cite{zhang2017deep}:} 
At the $i^\text{th}$ time interval, the inflow and outflow of the region $(h,w)$ are defined as follows:
$x_i^{in,h,w} = \sum \limits_{T_r \in \mathbb{P}} |\{ j>1 | g_{j-1}\notin(h,w)~\&~g_j \in (h,w)\}|, 
x_i^{out,h,w} = \sum \limits_{T_r \in \mathbb{P}} |\{ j>1 | g_{j-1} \in (h,w)~\&~g_j \notin (h,w)\}|,$
where $\mathbb{P}$ is a set of trajectories at the $i^\text{th}$ time interval.
For each trajectory $T_r: g_1 \rightarrow g_2 \rightarrow \cdots \rightarrow g_{T_r}$ in the trajectory set $\mathbb{P}$, $g_j \in (h,w)$ indicates that location $g_j$ lies insider the region $(h,w)$ , and vice versa.
Tensor $\mathbf{X}_{i} \in  \mathbb{R}^{2 \times H \times W}$ ($(\mathbf{X}_{i})_{0, h, w}=x_i^{in,h,w}$ and $(\mathbf{X}_{i})_{1, h, w}=x_i^{out,h,w}$) can be used to represent the urban flow of the whole city are.
For simplification, in this paper, we use $\mathbf{X}_{t}^c$, $\mathbf{X}_{t}^p$, and $\mathbf{X}_{t}^q$ to represent the urban flow tensor in the \textit{closeness}, \textit{period}, and \textit{trend} that correspond to the recent time intervals, daily periodicity, and weekly trend.
Specifically, for the same $t$ value, 
the time in between $\mathbf{X}_{t}^c$ and $\mathbf{X}_{t}^p$ is a day apart, whereas the time in between $\mathbf{X}_{t}^c$ and $\mathbf{X}_{t}^q$ is a week apart (see Fig.~\ref{fig:intro} (c)).

\noindent\textbf{Flow Prediction:} 
Given \textit{trend} $\{\mathbf{X}_{1}^q, \mathbf{X}_{2}^q, \cdots, \mathbf{X}_{T}^q\}$, \textit{period} $\{\mathbf{X}_{1}^p, \mathbf{X}_{2}^p, \cdots, \mathbf{X}_{T}^p\}$, and \textit{closeness} urban flow tensors $\{\mathbf{X}_{1}^c, \mathbf{X}_{2}^c, \cdots, \mathbf{X}_{T-1}^c\}$,
the goal is to predict the flow in the next time interval, i.e, $\mathbf{\hat{X}}_{T}^c$. 
Here, $T-1$ is the observation length.

\section{Methodology}
The architecture of our method for urban flow prediction is illustrated in Fig.~\ref{fig:framework}.
It consists of three major modules: 
% \begin{itemize}
    (i) \textit{Short-Term Prediction} module (the light green part in Fig.~\ref{fig:framework}) to yield a preliminary prediction based on the recent observations, i.e., the \textit{closeness} component.
    (ii) A \textit{Long-Term Relation Prediction} module (the light yellow part in Fig.~\ref{fig:framework}) to model the long-term relation per time step based on the \textit{period} component (daily periodicity) and the \textit{trend} component (weekly periodicity).
    Such long-term periodic influence is then incorporated into the urban flow prediction.
    (iii) A \textit{Prediction Consistency} module (the light blue part in Fig.~\ref{fig:framework}) to measure the consistency between the inferred relation calculated from the predicted $\mathbf{\hat{X}}_{T}^c$ and the \pr\ from the above long term relation prediction module.
% \end{itemize}

\subsection{Short-Term Prediction} \label{sec:short}
Intuitively, the most up-to-date data plays an important role in the prediction of the next time interval urban flow.
This module focuses on the short-term pattern with the recent time intervals flow tensors, i.e., the \textit{closeness} component.
With tensors $\{\mathbf{X}_{1}^c, \mathbf{X}_{2}^c, \cdots, \mathbf{X}_{T-1}^c\}$ as input, we can get an initial \ptf\ $\mathbf{\tilde{X}}_{T}^{c}$ which serves as the foundation of the final prediction $\mathbf{\hat{X}}_{T}^c$.
This process is modeled as:
\begin{equation}
    \mathbf{\tilde{X}}_{T}^{c} = \mathcal{F} (\mathbf{X}_{1}^c, \mathbf{X}_{2}^c, \cdots, \mathbf{X}_{T-1}^c),
\end{equation}
where $\mathcal{F}(\cdot)$ represents the prediction function (the Short-Term Prediction Unit that is shown as the green box in Fig.~\ref{fig:framework}).
% It could be any sequence prediction method such as LSTM.
In the proposed~\name, the widely-used \textit{Residual Unit} proposed by ST-ResNet~\cite{zhang2017deep} is selected as the base model for the Short-Term Prediction Unit.
% $\mathcal{F}(\cdot)$.
It has shown a good ability to model the spatial dependencies in the city scale.
Note that using the \textit{Residual Unit} is the only similarity between our~\name\ and ST-ResNet.

\subsection{Long-Term Relation Prediction} \label{sec:long}
In this module, we focus on the daily periodicity and weekly trend and exploring how to model the long-term pattern for urban flow prediction.
Unlike existing methods that process the \textit{period} and \textit{trend} components individually, we simultaneously process these components and extract the periodic relation.
For example, what is the relation between the urban of 8 pm today (Friday), 8 pm yesterday (Thursday), and 8 pm last week's Friday?
Assuming that our predicting target is the urban flow of 9 pm today, our aim is to use the relations of previous time intervals (e.g., 5 pm and 8 pm) to facilitate and improve the performance.

For $1 \leq t \leq T-1$, all three urban flow tensors ($\mathbf{X}_{t}^{c}$ from the \textit{closeness} component, $\mathbf{X}_{t}^{p}$ from the \textit{period} component, and $\mathbf{X}_{t}^{q}$ from the \textit{trend} component) are known.
To model the relation of these three tensors, similar to~\cite{santoro2017simple}, a multi-layer perceptrons (MLP) $g(\cdot)$ (the Long-Term Relation Unit which is shown as the yellow box in Fig.~\ref{fig:framework}) is used to learn the relation vector $\mathbf{R}_{t}$:
\begin{equation}
    \mathbf{R}_{t} = g(\mathbf{X}_{t}^{c} \oplus \mathbf{X}_{t}^{p} \oplus \mathbf{X}_{t}^{q}), \label{eq:relation}
\end{equation}
where $\oplus$ is the concatenation operation.
Given the sequence of relation vectors from $t=1$ to $t=T-1$, the relation for $t=T$ can be predicted through:
\begin{equation}
    \mathbf{\tilde{R}}_{T} = \text{Transformer}(\mathbf{R}_{1}, \mathbf{R}_{2}, \cdots, \mathbf{R}_{T-1}). \label{eq:transformer}
\end{equation}
Here, we take the merit of the \textit{Transformer}~\cite{vaswani2017attention}
% \footnote{In our work, the standard implementation of Transformer from PyTorch (version 1.5.0) is used.}, 
a popular architecture for sequence prediction.
The self-attention mechanisms of the transformer is suitable for modeling pairwise interactions between two relation vectors of two time intervals
% (e.g., 5 pm and 8 pm, see the example of sequence $m$ in Fig.~\ref{fig:position_encoding}) 
in the input sequence.
Based on the \pr\ vector $\mathbf{\tilde{R}}_{T}$, a tensor $\mathbf{\tilde{X}}_{T}^{r} \in \mathbb{R}^{2 \times H \times W}$ is generated via:
\begin{equation}
       \mathbf{\tilde{X}}_{T}^{r} = \text{MLP}_{\text{R}}(\mathbf{\tilde{R}}_{T}), \label{eq:decode_relation}
\end{equation}
where $\text{MLP}_{\text{R}}(\cdot)$ stands for a MLP architecture.

The tensor $\mathbf{\tilde{X}}_{T}^{r}$ can be seen as a supplement to the initial prediction $\mathbf{\tilde{X}}_{T}^{c}$ from the short-term prediction module.
The rationale of this long-term relation prediction process is that given the \pr, we should be able to ``decode'' a predicted tensor.
This decoded tensor helps the prediction of $\mathbf{\hat{X}}_{T}^c$ as the relation vector $\mathbf{\tilde{R}}_{T}$ models the long-term periodicity at $T$.

% \subsubsection{Position Encoding} \label{sec:pe}
\smallskip
\noindent\textbf{Position Encoding.}
Given that there is no recurrence operation in the Transformer architecture, position encodings are often added to each input vector of the Transformer to compensate the missing sequential order information~\cite{vaswani2017attention}.
These position encodings are modeled as sine and cosine functions:
\begin{align}
\label{eq:pe_1}
    PE_{(pos, 2k)} &= \sin (pos/10000^{2k/d_{\textbf{R}}}), \\ \label{eq:pe_2}
    PE_{(pos, 2k+1)} &= \cos (pos/10000^{2k/d_{\textbf{R}}}), 
\end{align}
where $pos$ is the position index and $k$ is the dimension index.
$d_{\textbf{R}}$ is the dimension of the relation vector $\mathbf{R}_{t}$ (the input of the Transformer~\name).

Unlike other Transformer based applications such as translation~\cite{vaswani2017attention} and object detection~\cite{carion2020end}, in the proposed~\name, relation vector at each time interval (the input sequence of Eq.~\eqref{eq:transformer}) has its own unique periodic position to indicate the absolute temporal position in one day.
Thus, we explicitly design a periodic position encoding strategy to model the periodic position information.
Since it represents the absolute position in one day of each time interval, the maximum $pos$ value in Eq.~\eqref{eq:pe_1} and Eq.~\eqref{eq:pe_2} depends on the total number of time intervals in one day (e.g, 24 for 1 hour interval and 48 for 0.5 hour interval).

\subsection{Prediction with Consistency}

In the urban flow prediction task, extra information such as the date information and time in the day (e.g., the prediction target is \textit{11 am} on \textit{Friday} in the example shown in Fig.~\ref{fig:intro}) also have influences on the prediction.
In~\name, such influence is modeled as a tensor $\mathbf{\tilde{X}}_{T}^{\text{extra}} \in \mathbb{R}^{2 \times H \times W}$ through a MLP ($\text{MLP}_{\text{extra}}$ in Fig.~\ref{fig:framework}) with the input of the extra information $\mathbf{X}_{T}^{\text{extra}}$ at the prediction target $T$.

\begin{equation}
    \mathbf{\hat{X}}_{T}^{c} =  \mathbf{W}_1 \circ \mathbf{\tilde{X}}_{T}^{c} + \mathbf{W}_2 \circ \mathbf{\tilde{X}}_{T}^{r} + \mathbf{W}_3 \circ \mathbf{\tilde{X}}^{\text{extra}}_{T} \label{eq:out}
\end{equation}
As given in Eq.~\eqref{eq:out} where $\circ$ represents the element-wise multiplication, the final predicted urban flow tensor $\mathbf{\hat{X}}_{T}^{c}$ is a weighted fusion of three parts:
(1)~the initial predicted $\mathbf{\tilde{X}}_{T}^{c}$ from the short-term prediction;
(2)~the $\mathbf{\tilde{X}}_{T}^{r}$ based on the long-term relation modeling;
and (3)~the influence $\mathbf{\tilde{X}}^{\text{extra}}_{T}$ from the extra information.
More detailed discussion of this weighted combination step is given in Sec.~\ref{sec:weight}.

Given two known history observations $\mathbf{X}_{T}^p$ and $\mathbf{X}_{T}^q$, we can infer a relation based on the predicted $\mathbf{\hat{X}}_{T}^{c}$ by:
\begin{equation}
    \mathbf{\hat{R}}_{T} = g(\mathbf{\hat{X}}_{T}^{c} \oplus \mathbf{X}_{T}^{p} \oplus \mathbf{X}_{T}^{q})
\end{equation}
Intuitively, this inferred relation $\mathbf{\hat{R}}_{T}$ should be consistent with the \pr\ $\mathbf{\tilde{R}}_{T}$ given by Eq.~\eqref{eq:transformer}.
To further build a connection between these two vectors, we propose to use the cosine similarly as a scoring function to model the consistency between the inferred relation vector $\mathbf{\hat{R}}_{T}$ and the \pr\ vector $\mathbf{\tilde{R}}_{T}$ at $T$.

\smallskip
\noindent\textbf{Loss Function.}
Unlike existing urban flow prediction methods that only use the mean squared error between the predicted flow and the ground truth flow as the loss function,
the loss function in our~\name\ contains two terms:
\begin{equation}
     \mathcal{L} = \alpha \| \mathbf{X}_{T}^{c} - \mathbf{\hat{X}}_{T}^{c} \|_{2}^{2} + \beta (1 - \frac{{\mathbf{\hat{R}}_{T}}\cdot {\mathbf{\tilde{R}}_{T}} }{\|\mathbf{\hat{R}}_{T} \| \| \mathbf{\tilde{R}}_{T}\|}), \label{eq:loss}
\end{equation}
where $\alpha$ and $\beta$ are the weights of two terms.
The first term is the widely used mean squared error and the second term is the above cosine similarly based consistency loss.
It gives more penalty if the inferred relation and the \pr\ are not consistent.

\section{Experiments}

\subsection{Datasets and Metrics}
In this paper, we focus on spatio-temporal raster urban flow data and conduct experiments on the following three real-world urban flow datasets:
(1)~BikeNYC~\cite{lin2019deepstn}: This dataset is extracted from the New York City Bike system from the time period: 2014-04-01 to 2014-09-30.
(2)~TaxiNYC~\cite{yao2019revisiting}: This datasets consists of taxi trip records of New York City from the period: 2015-01-01 to 2015-03-01.
(3)~TaxiBJ~\cite{zhang2017deep}: This dataset is the taxicab GPS data collected in Beijing from time periods: 2013-07-01 to 2013-10-30, 2014-05-01 to 2014-06-30, 2015-03-01 to 2015-06-30, and 2015-11-01 to 2016-04-10.
Since there is no connectivity information such as road maps, graph based spatio-temporal traffic prediction models are irrelevant.
% Detailed descriptions of these datasets are listed in Table~\ref{tab:data}.
For each dataset, the first 80\% data is used for training and the rest 20\% data is used for testing.
The Min-Max normalization is adopted to transform the urban flow values into the range $[0, 1]$.
During the evaluation process, we rescale predicted values back to the normal values to compare with the ground truth data.
We evaluate our method with two commonly used metrics: the Rooted Mean Squared Error (RMSE) and the Mean Absolute Error (MAE).

\subsection{Implementation details}
The hyperparameters are set based on the performance on the validation set (10\% of the training data).
The Adam optimizer is used to train our models.
The base unit used in our short-term prediction module consists of three $3\times 3$ convolutional layers with 32 filters and one $3\times 3$ convolutional layer with 2 filters (corresponding to the in and out flow in each urban flow tensor).
The dimension of each relation vector is 256, i.e., $\mathbf{R}_{t}\in \mathbb{R}^{256}$.
As for the loss function~\eqref{eq:loss}, both weights $\alpha$ and $\beta$ are set to 1.
Following VLUC-Net~\cite{jiang2019vluc}, we set the observation length to 6.
% $T=7$ which means  is 6.
That is, the length of the \textit{closeness} component is 6, whereas the length of the \textit{period} and \textit{trend} components are 7.
For the extra information, considering that information such as weather is not available for all three datasets, only the temporal metadata (i.e., the time interval of a day and the day of a week) is used.
The reported results of our~\name\ and its variants in the experiments are the average of 5 runs.

\subsection{Comparison against other methods}
In our experiments, we compare \name\ against the following urban flow prediction methods: HistoricalAverage (HA); Convolutional LSTM (ConvLSTM); ST-ResNet \cite{zhang2017deep}; DMVST \cite{yao2018deep}; DeepSTN+~\cite{lin2019deepstn}; STDN~\cite{yao2019revisiting}; VLUC-Net~\cite{jiang2019vluc}.

The upper half of Table~\ref{tab:results} shows the results of the proposed~\name\ as compared to other methods on three datasets. 
For each column, the best result is given in bold.
Specifically, compared to other deep learning based prediction methods, the HA has the worst performance as it can only make predictions by simply averaging historical observations.
In general, \name\ achieves the best performance on all three public datasets for both RMSE and MAE.
STDN and VLUC-Net share the second best performance depending on the dataset and the metric.
Compared to our~\name, in other methods, the three components are connected merely through attention based fusion, which overlooks the periodic relation between the \textit{closeness}, \textit{period}, and \textit{trend} components.
These comparison results demonstrate the superior of the proposed method. 

\subsection{Ablation studies}
To explore the effectiveness of each module in our proposed method, we consider the following three variants:
% \begin{itemize}
    (i) V1: We discard the short-term prediction module for this variant. That is, the \textit{closeness} component is not used in the prediction process.
    (ii) V2: The long-term relation module is removed. Since the consistency loss is based on the long-term relation module, it is disabled in this variant as well.
    (iii) V3: For this variant, we only remove the consistency loss in~\name, which means the second term (used to measure the cosine similarity between the inferred relation and the \pr\ in Eq.~\eqref{eq:loss}) is dropped.

The results of these variants on three datasets are presented in the lower half of Table~\ref{tab:results}.
We also include the results of our~\name\ in the table for comparison.
In general, V1 has the worst results in these variants.
Actually, it is only better than HA in Table~\ref{tab:results}.
This is as expected because the \textit{closeness}  component contains the most up-to-date urban flow information for prediction.
We found that the usage of only historical data (i.e., the \textit{period} component of the previous day and the \textit{trend} component of the previous week) is not enough for accurate urban flow prediction.

\begin{table}[!t]
\centering
\caption{The RMSE / MAE results on three datasets. The upper half and the latter half list the performance of existing methods and variants of our~\name, respectively.}
\label{tab:results}
\addtolength{\tabcolsep}{0.5ex}
\begin{tabular}{|l||c|c|c|c|c|c|} \hline
 & \multicolumn{2}{c|}{BikeNYC} & \multicolumn{2}{c|}{TaxiNYC} & \multicolumn{2}{c|}{TaxiBI} \\ \cline{2-7}
 & RMSE & MAE & RMSE & MAE & RMSE & MAE \\\hline
HA & 15.676 & 4.882 & 21.535 & 7.121 & 45.004 & 24.475 \\
ConvLSTM & 6.616 & 2.412 & 12.143 & 4.811 & 19.247 & 10.816 \\
ST-ResNet & 6.106 & 2.360 & 11.553 & 4.535 & 18.702 & 10.493 \\
DMVST & 7.990 & 2.833 & 13.605 & 4.928 & 20.389 & 11.832 \\
DeepSTN+ & 6.205 & 2.489 & 11.420 & 4.441 & 18.141 & 10.126 \\
STDN & 5.783 & 2.410 & 11.252 & 4.474 & 17.826 & 9.901 \\
VLUC-Net & 5.831 & 2.175 & 10.654 & 4.157 & 18.378 & 10.325 \\\hline\hline
V1 & 10.743 & 3.462 & 19.292 & 6.298 & 23.529 & 13.273 \\
V2 & 6.331 & 2.462 & 11.726 & 4.816 & 18.998 & 11.197 \\
V3 & 5.971 & 2.215 & 10.695 & 4.077 & 17.442 & 10.068 \\\hline
\name & \textbf{5.729} & \textbf{2.139} & \textbf{10.395} & \textbf{3.955} & \textbf{17.017} & \textbf{9.778} \\ \hline
\end{tabular}
\end{table}

The prediction performance of V2 leads V1 by a relatively large margin but is worse than V3 in which the long-term prediction module is enabled.
Furthermore, if we compare V2 against other methods reported in Table~\ref{tab:results}, it can be noticed that the performance of V2 is close to ST-ResNet. 
Since the Residual Unit from ST-ResNet is used as the forecasting function in the short-term prediction module to predict urban flow based on the \textit{closeness}  component, it can be seen as a simplified version of ST-ResNet where the \textit{period} and \textit{trend} components are disabled.
Results of V3 outperform the other two variants but are worse than \name\ where the consistency loss is applied.
The proposed consistency loss can add more constraints during the training and stabilize the training process.
In a nutshell, it harmonizes the relation inferred from the \ptf\ and the \pr\ at the same time interval so that the predicted urban flow is more realistic.
Comparing the results from the three variants against \name, we observe that \name\ outperforms all the variants on all three datasets.
It demonstrates the effectiveness of each module and justifies the need for all modules to be included in our~\name.

\begin{table}
\centering
\caption{Six different configurations of the weighted fusion. C0 is our default configuration used in other experiments.}
\label{tab:weight}
\addtolength{\tabcolsep}{0.5ex}
\begin{tabular}{|c|ccc|}
\hline 
configurations & $\mathbf{W}_1$ & $\mathbf{W}_2$ & $\mathbf{W}_3$ \\ \hline\hline
C0 & \checkmark & \checkmark  & \checkmark  \\ \hline
C1 & \checkmark & \checkmark & $\times$ \\ \hline
C2 & \checkmark & $\times$ & \checkmark \\ \hline
C3 & $\times$ & \checkmark & \checkmark \\ \hline
C4 & $\times$ & $\times$ & $\times$ \\ \hline
C5 & \checkmark (softmax)  & \checkmark (softmax) & \checkmark (softmax) \\ \hline
\end{tabular}
\end{table}

\subsection{Different Weighted Fusion configurations}\label{sec:weight}

\begin{figure*}[!t]
    \centering
    \subfigure[BikeNYC: RMSE]{
    \includegraphics[width=.265\textwidth]{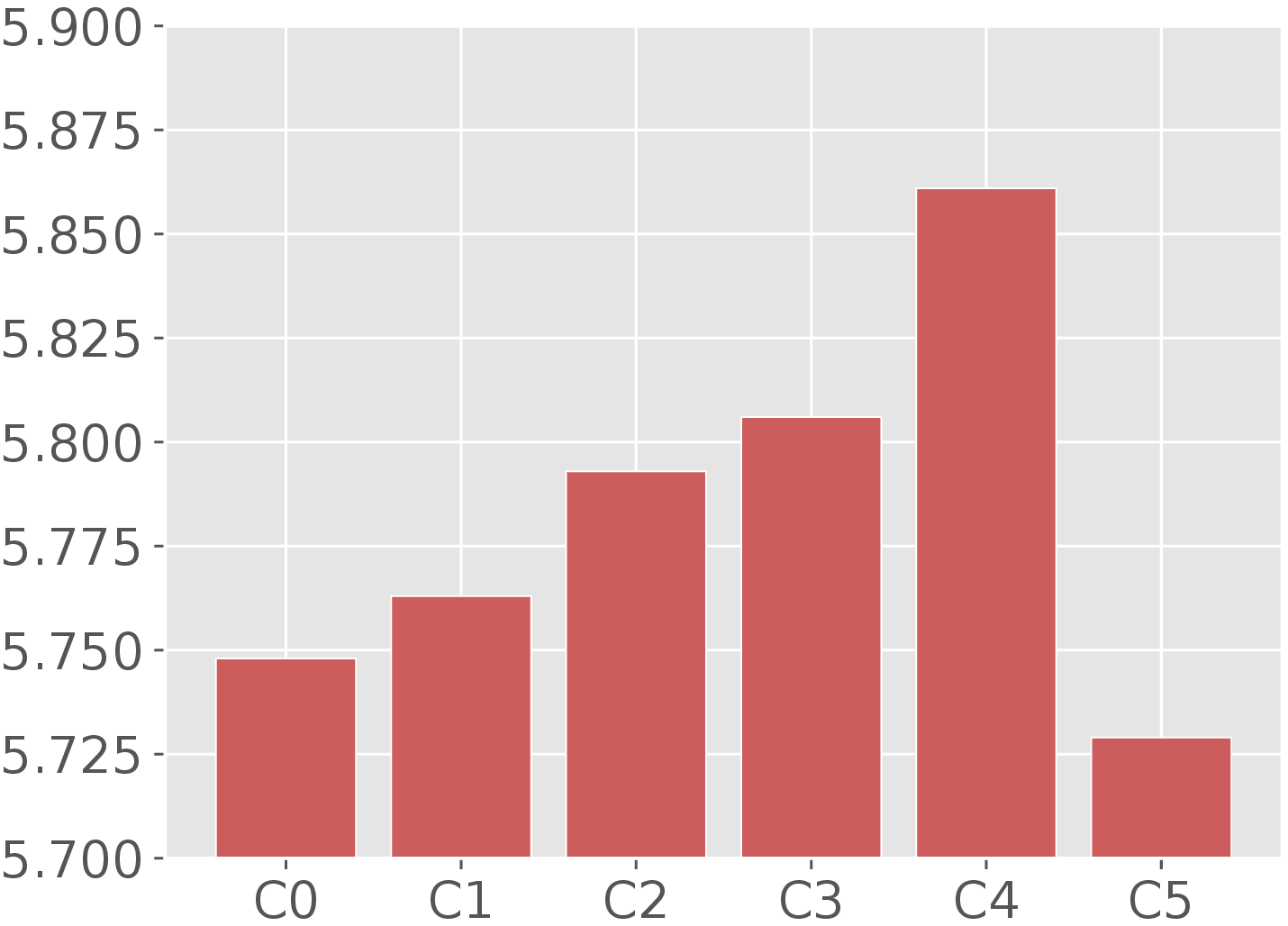}}
    \subfigure[TaxiNYC: RMSE]{
    \includegraphics[width=.265\textwidth]{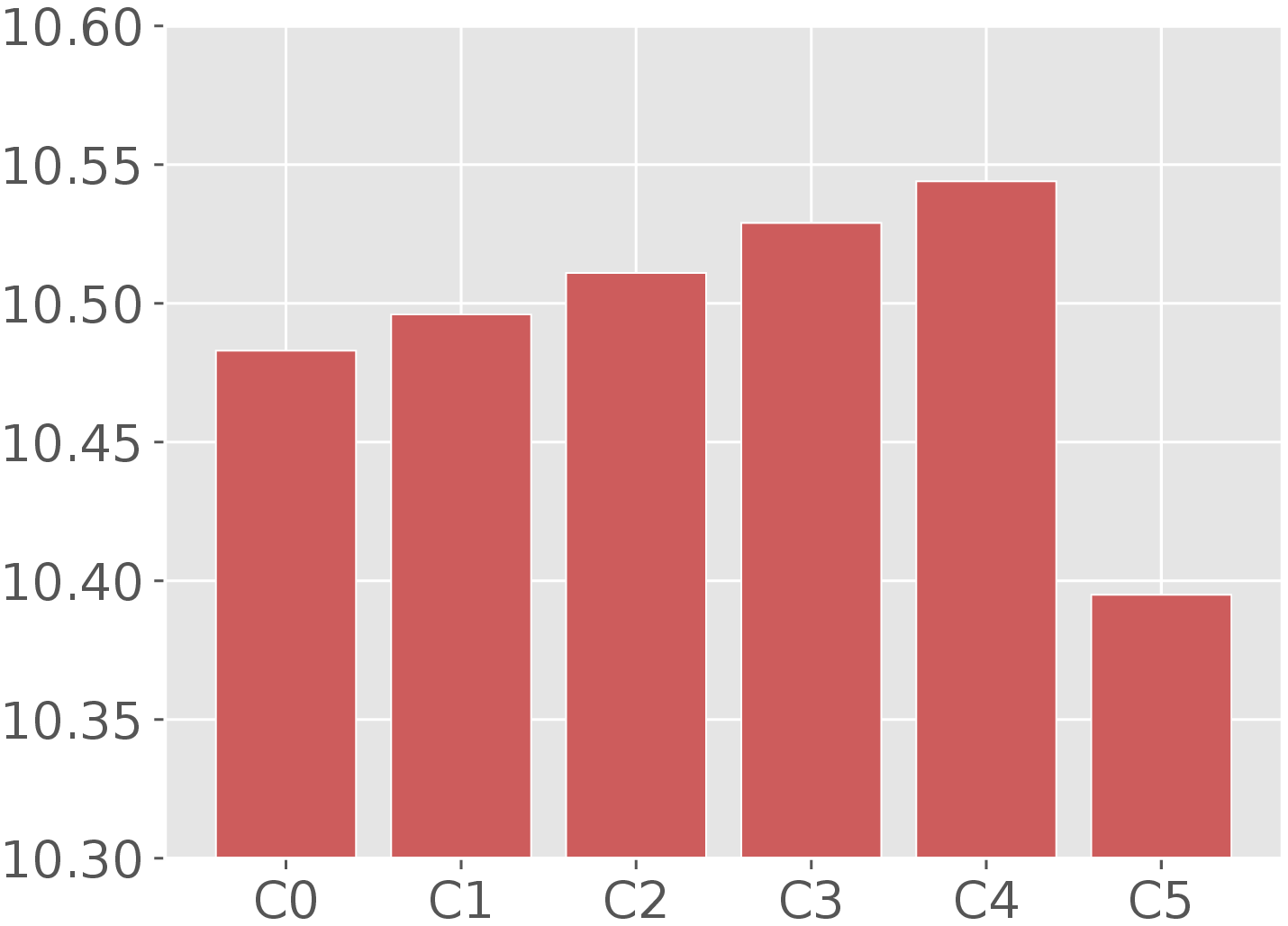}}
    \subfigure[TaxiBJ: RMSE]{
    \includegraphics[width=.265\textwidth]{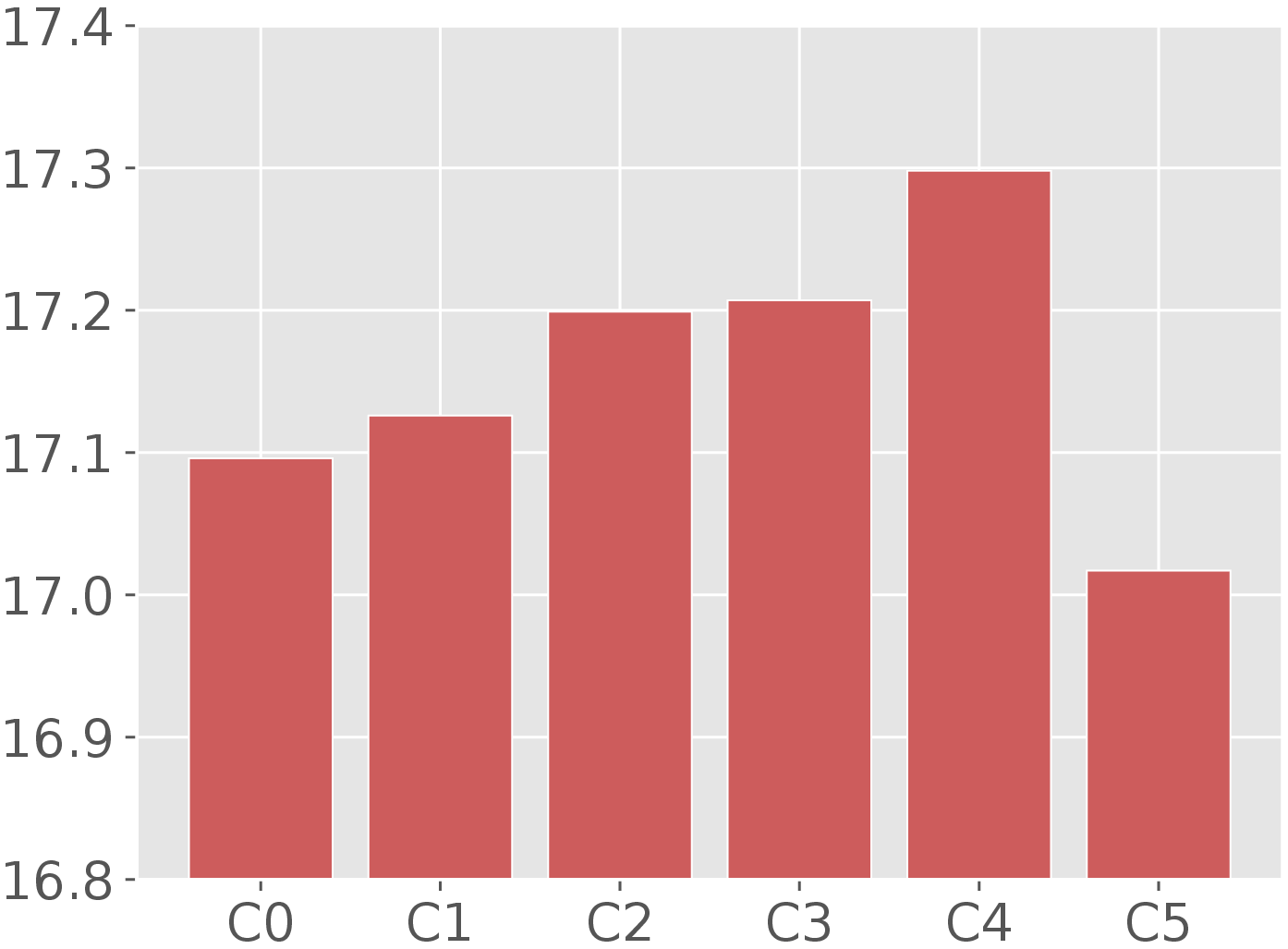}} \\
    \subfigure[BikeNYC: MAE]{
    \includegraphics[width=.265\textwidth]{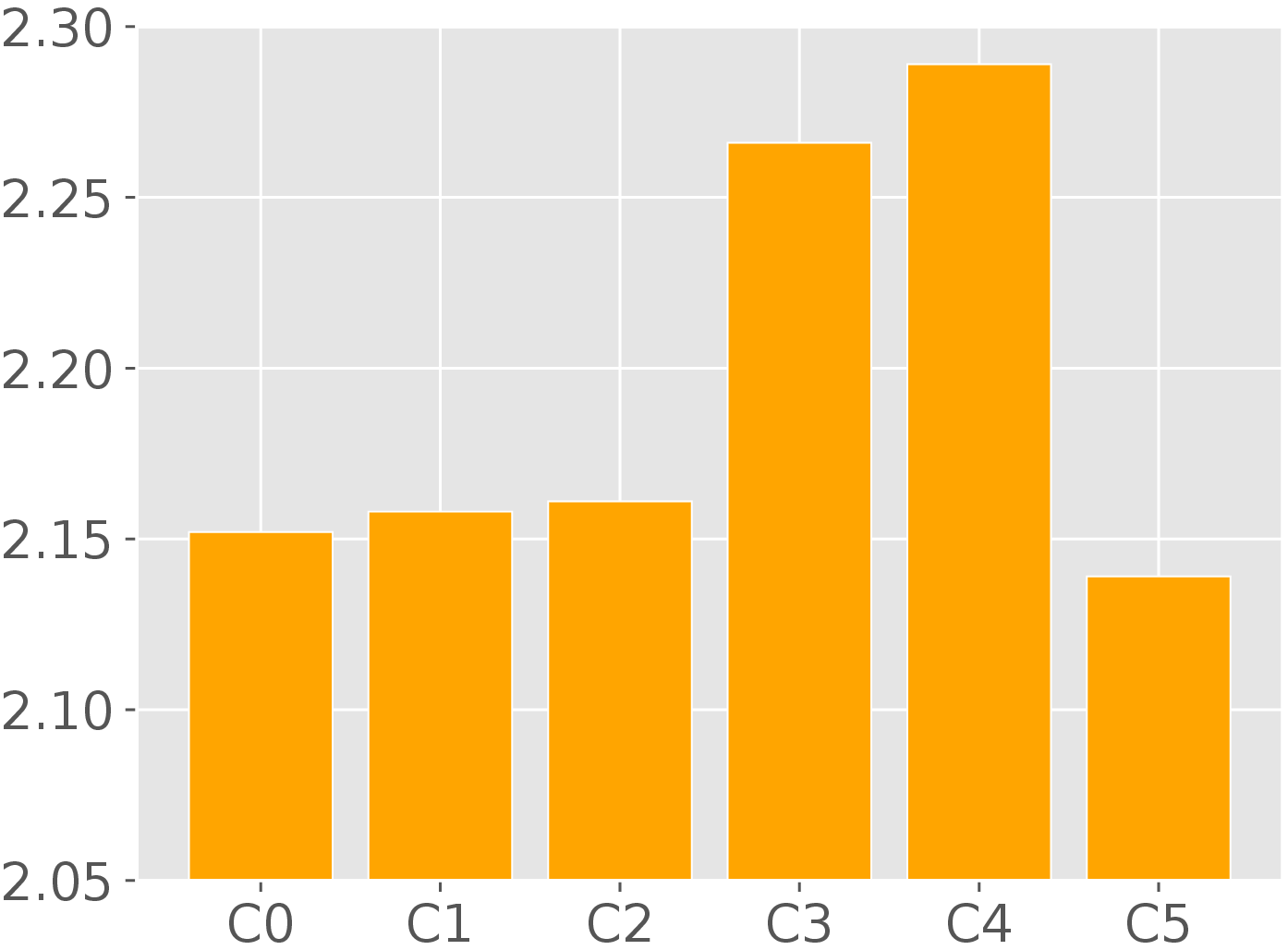}}
    \subfigure[TaxiNYC: MAE]{
    \includegraphics[width=.265\textwidth]{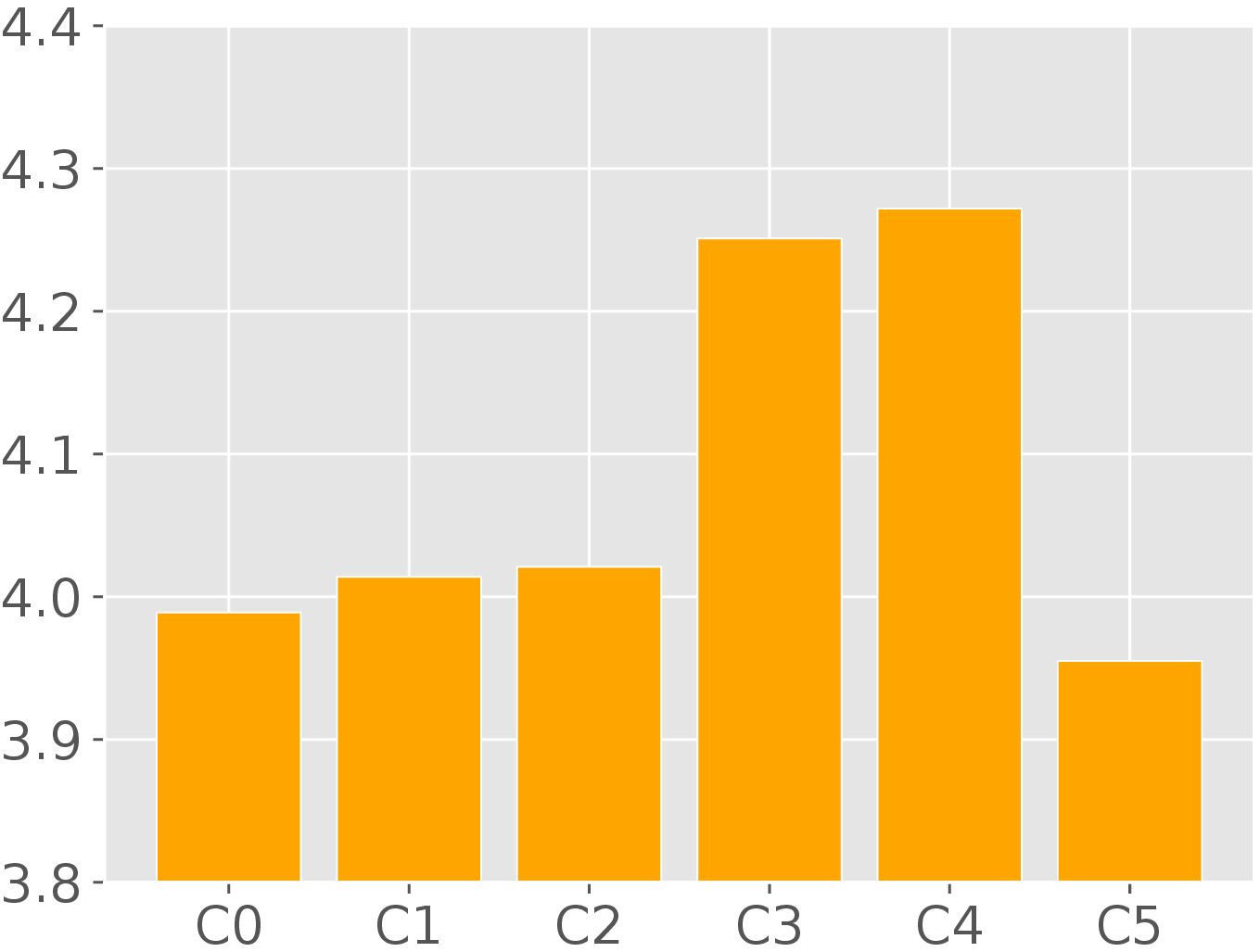}}
    \subfigure[TaxiBJ: MAE]{
    \includegraphics[width=.265\textwidth]{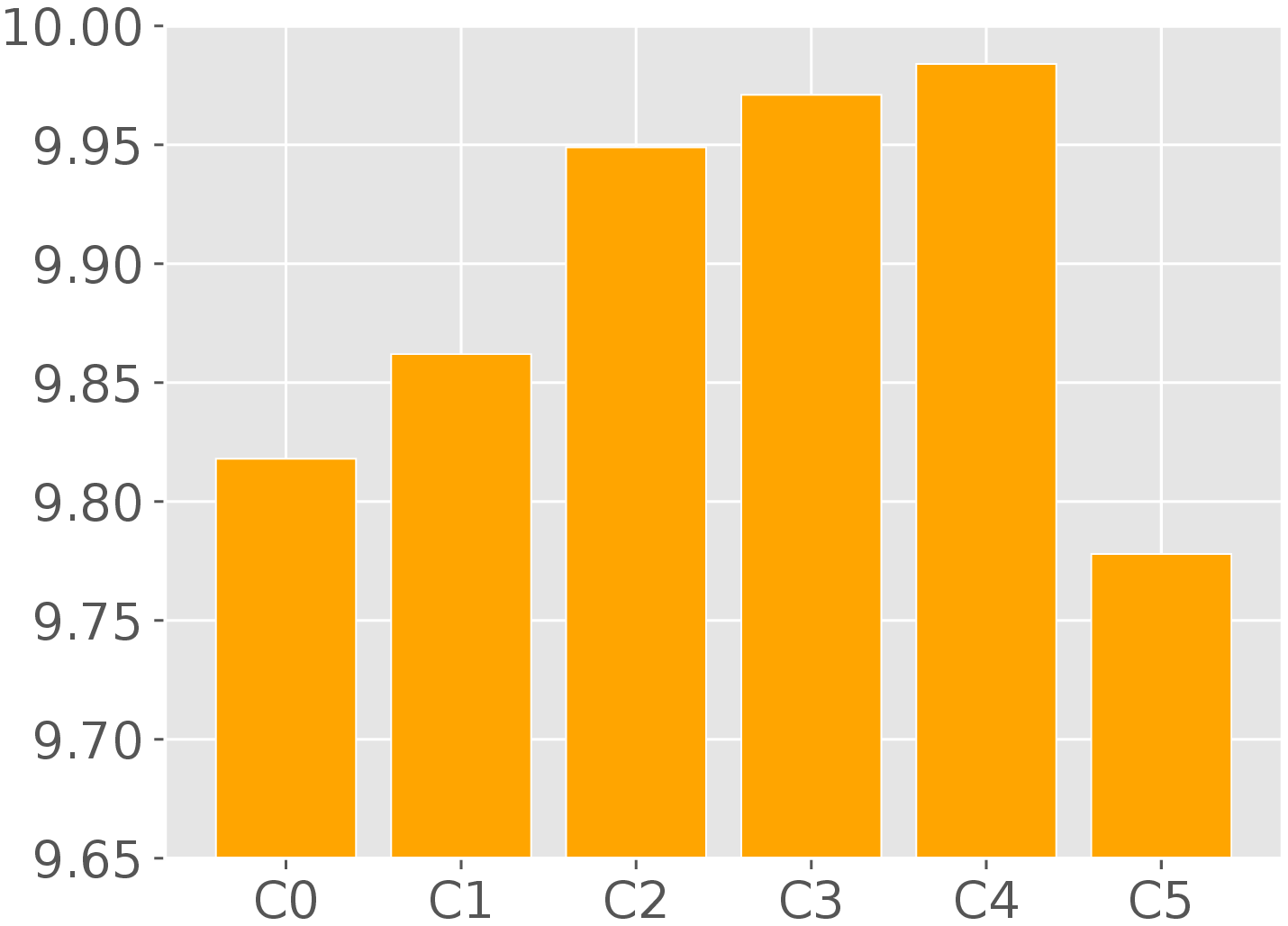}}
    \caption{The RMSE and MAE results of different configurations listed in Table~\ref{tab:weight}.}
    \label{fig:weight}
\end{figure*}

The weighted fusion (Eq.~\eqref{eq:out}) is an important step in \name\ as it combines the initial predicted $\mathbf{\tilde{X}}_{T}^{c}$ with the influence from the long-term relation and extra information.
To fully research this fusion step, six different fusion configurations are explored.
Table~\ref{tab:weight} summarizes the setting of each configuration.
A $\checkmark$ indicates that the corresponding weight matrix is enabled whereas a $\times$ means that the corresponding weight matrix is removed from Eq.~\eqref{eq:out}.
Taken the configuration C4 as an example, without all three weighting terms, the weighted fusion prediction step given in Eq.~\eqref{eq:out} will be degraded to a plain summation operation, i.e., $\mathbf{\hat{X}}_{T}^{c} =  \mathbf{\tilde{X}}_{T}^{c} + \mathbf{\tilde{X}}_{T}^{r} + \mathbf{\tilde{X}}^{\text{extra}}_{T}$.
Note that the last row (C5) in the table is the default configuration of~\name. The softmax function is applied to make sure that the sum of three weights with the same index in the three weight matrices equals to 1.
% Note that this configuration is used in other ex
C0 can be seen as a simplified configuration of C5.
% In the configuration C5, the softmax function is applied to make sure that the sum of three weights with the same index in the three weight matrices equals to 1.
The prediction results of these different fusion configurations are demonstrated in Fig.~\ref{fig:weight}.
% In the figure, the first row shows the RMSE results on three datasets whereas the second row gives the MAE results.

For configurations (C1-C4) where one or more weight matrices are disabled, their prediction performance are worse than C0.
It indicates that each weight matrix is necessary for the fusion step.
To be more specific, C4 has the worst results on all three datasets as it simply adds the periodic relation influence $\mathbf{\tilde{X}}_{T}^{r}$ and the extra influence $\mathbf{\tilde{X}}^{\text{extra}}_{T}$ to the initial predicted $\mathbf{\tilde{X}}_{T}^{c}$.
If only one weight matrix is removed (C1, C2, and C3), C3 cannot yield predictions as accurate as of the other two configurations.
C3 discards $\mathbf{W}_1$ which is the weight matrix of the initial predicted $\mathbf{\tilde{X}}_{T}^{c}$ generated through the short-term prediction module.
This result is consistent with the above ablation results that V1 does not perform well.
It further confirms that the short-term prediction module is a significant part of the traffic flow prediction. 
For the configuration C5, its performance stands out in the figure and outperforms all the other configurations.
It reveals that applying an extra softmax operation on these weight matrices helps to further improve the traffic flow prediction performance.

\section{Conclusion}
In this paper, we present a novel method called \name\ for the urban flow prediction task.
We explicitly design a long-term relation prediction module to better capture the periodicity in the urban flow sequence.
Through the prediction consistency module and the consistency loss, \name\ is able to make the \ptf\ and the \pr\ be consistent with each other so that the prediction performance can be improved further.
We conduct extensive experiments on three real-world datasets.
The experimental results demonstrate that the proposed \name\ outperforms other prediction methods and show the effectiveness of each module in \name.
In addition, we explore and compare different weighted fusion methods in our experiments.
These studies and findings would be of interest to other researchers in related areas.

\subsubsection*{Acknowledgments.}
We acknowledge the support of Australian Research Council
Discovery Project \textit{DP190101485}.

\bibliographystyle{splncs04}
\bibliography{main}

\end{document}